\documentclass[11pt,a4paper]{article}
\usepackage[hyperref]{acl2020}
\usepackage{times}
\usepackage{latexsym}

\usepackage{microtype}
\usepackage{graphicx}
\usepackage[ruled]{algorithm2e}
\usepackage{booktabs}
\usepackage{multirow}

\aclfinalcopy 

\renewcommand{\mbox}{}
\newcommand{\BluebertC}{\mbox{BlueBERT\textsubscript{clinical}}\xspace}
\newcommand{\MTBluebertCR}{\mbox{MT-BlueBERT-Refinement\textsubscript{clinical}}\xspace}
\newcommand{\MTBluebertCFT}{\mbox{MT-BlueBERT-Fine-Tune\textsubscript{clinical}}\xspace}
\newcommand{\BluebertB}{\mbox{BlueBERT\textsubscript{biomedical}}\xspace}
\newcommand{\MTBluebertBR}{\mbox{MT-BlueBERT-Refinement\textsubscript{biomedical}}\xspace}
\newcommand{\MTBluebertBFT}{\mbox{MT-BlueBERT-Fine-Tune\textsubscript{biomedical}}\xspace}
\newcommand{\MTClinicalBERTFT}{\mbox{MT-ClinicalBERT-Fine-Tune}\xspace}
\newcommand{\MTBioBERTFT}{\mbox{MT-BioBERT-Fine-Tune}\xspace}
\newcommand{\MTBERT}{\mbox{MT-BERT}\xspace}
\newcommand{\MTBERTFT}{\mbox{MT-BERT-Fine-Tune}\xspace}
\newcommand{\MTBERTR}{\mbox{MT-BERT-Refinement}\xspace}

\title{An Empirical Study of Multi-Task Learning on BERT \\for Biomedical Text Mining}

\author{Yifan Peng \hspace{1em} Qingyu Chen \hspace{1em} Zhiyong Lu \\
  National Center for Biotechnology Information \\
  National Library of Medicine, National Institutes of Health \\
  Bethesda, MD, USA \\
  \texttt{\{yifan.peng, qingyu.chen, zhiyong.lu\}@nih.gov}}

\date{}

\begin{document}
\maketitle
\begin{abstract}
Multi-task learning (MTL) has achieved remarkable success in natural language processing applications. In this work, we study a multi-task learning model with multiple decoders on varieties of biomedical and clinical natural language processing tasks such as text similarity, relation extraction, named entity recognition, and text inference. Our empirical results demonstrate that the MTL fine-tuned models outperform state-of-the-art transformer models (e.g., BERT and its variants) by 2.0\% and 1.3\% in biomedical and clinical domains, respectively. Pairwise MTL further demonstrates more details about which tasks can improve or decrease others. This is particularly helpful in the context that researchers are in the hassle of choosing a suitable model for new problems. The code and models are publicly available at \url{https://github.com/ncbi-nlp/bluebert}.
\end{abstract}

\section{Introduction}

Multi-task learning (MTL) is a field of machine learning where multiple tasks are learned in parallel while using a shared representation~\cite{caruana1997multitask}. Compared with learning multiple tasks individually, this joint learning effectively increases the sample size for training the model, thus leads to performance improvement by increasing the generalization of the model~\cite{zhang2017survey}. This is particularly helpful in some applications such as medical informatics where (labeled) datasets are hard to collect to fulfill the data-hungry needs of deep learning.

MTL has long been studied in machine learning~\cite{ruder2017overview} and has been used successfully across different applications, from natural language processing~\cite{collobert2008unified, luong2015multi, liu2019multi}, computer vision~\cite{wang2009boosted, liu2019end, chen2019multi}, to health informatics~\cite{zhou2011multi, he2016novel, harutyunyan2019multitask}. MTL has also been studied in biomedical and clinical natural language processing (NLP) such as named entity recognition and normalization and the relation extraction. However, most of these studies focus on either one task with multi corpora~\cite{khan2020mt, wang2019cross} or multi-tasks on a single corpus~\cite{xue2019fine, li2017neural, zhao2019neural}.

To bridge this gap, we investigate the use of MTL with transformer-based models (BERT) on multiple biomedical and clinical NLP tasks. We hypothesize the performance of the models on individual tasks (especially in the same domain) can be improved via joint learning. Specifically, we compare three models: the independent single-task model (BERT), the model refined via MTL (called \MTBERTR), and the model fine-tuned for each task using MT-BERT-Refinement (called \MTBERTFT). We conduct extensive empirical studies on the Biomedical Language Understanding Evaluation (BLUE) benchmark~\cite{peng2019transfer}, which offers a diverse range of text genres (biomedical and clinical text) and NLP tasks (such as text similarity, relation extraction, and named entity recognition). When learned and fine-tuned on biomedical and clinical domains separately, we find that MTL achieved over 2\% performance on average, created new state-of-the-art results on four BLUE benchmark tasks. We also demonstrate the use of multi-task learning to obtain a single model that still produces state-of-the-art performance on all tasks. This positive answer will be very helpful in the context that researchers are in the hassle of choosing a suitable model for new problems where training resources are limited. 

Our contribution in this work is three-fold: (1)~We conduct extensive empirical studies on 8 tasks from a diverse range of text genres. (2) We demonstrate that the MTL fine-tuned model (\MTBERTFT) achieved state-of-the-art performance on average and there is still a benefit to utilizing the MTL refinement model (\MTBERTR). Pairwise MTL, where two tasks were trained jointly, further demonstrates which tasks can improve or decrease other tasks. (3) We make codes and pre-trained MT models publicly available.

The rest of the paper is organized as follows. We first present related work in Section~\ref{sec:rel}. Then, we describe the multi-task learning in Section~\ref{sec:multitask}, followed by our experimental setup, results, and discussion in Section~\ref{sec:exp}. We conclude with future work in the last section.

\section{Related work}
\label{sec:rel}

Multi-tasking learning (MTL) aims to improve the learning of a model for task t by using the knowledge contained in the tasks where all or a subset of tasks are related~\cite{zhang2017survey}. It has long been studied and has applications on neural networks in the natural language processing domain~\cite{caruana1997multitask}. \citet{collobert2008unified} proposed to jointly learn six tasks such as part-of-speech tagging and language modeling in a time-decay neural network. \citet{changpinyo2018multi} summarized recent studies on applying MTL in sequence tagging tasks. \citet{bingel2017identifying} and \citet{martinezalonso2017when} focused on conditions under which MTL leads to gain in NLP, and suggest that certain data features such as learning curve and entropy distribution are probably better predictors of MTL gains. 

In the biomedical and clinical domains, MTL has been studied mostly in two directions. One is to apply MTL on a single task with multiple corpora. For example, many studies focused on named entity recognition (NER) tasks~\cite{crichton2017neural, wang2019multitask, wang2019cross}. \citet{zhang2018multitask}, \citet{khan2020mt}, and \citet{mehmood2019multi} integrated MTL in the transformer-based networks (BERT), which is the state-of-the-art language representation model and demonstrated promising results to extract biomedical entities from literature. \citet{yang2019information} extracted clinical named entity from Electronic Medical Records using LSTM-CRF based model. Besides NER, \citet{li2018multi} and \citet{li2019syntax} proposed to use MTL on relation classification task and \citet{du2017novel} on biomedical semantic indexing. \citet{xing2018adaptive} exploited domain-invariant knowledge to segment Chinese word in medical text.

The other direction is to apply MTL on different tasks, but the annotations are from a single corpus. \citet{li2017neural} proposed a joint model extract biomedical entities as well as their relations simultaneously and carried out experiments on either the adverse drug event corpus~\cite{gurulingappa2012development} or the bacteria biotope corpus~\cite{deleger2016overview}. \citet{shi2019family} also jointly extract entities and relations but focused on the BioCreative/OHNLP 2018 challenge regarding family history extraction~\cite{liu2018overview}. \citet{xue2019fine} integrated the BERT language model into joint learning through dynamic range attention mechanism and fine-tuned NER and relation extraction tasks jointly on one in-house dataset of coronary arteriography reports. 

Different from these works, we studied to jointly learn 8 different corpora from 4 different types of tasks. While MTL has brought significant improvements in medicine tasks, no (or mixed) results have been reported when pre-training MTL models in different tasks on different corpora. To this end, we deem that our model can provide more insights about conditions under which MTL leads to gains in BioNLP and clinical NLP, and sheds light on the specific task relations that can lead to gains from MTL models over single-task setups.

\section{Multi-task model}
\label{sec:multitask}

\begin{figure*}[ht]
\centering
\includegraphics[width=.9\textwidth,clip,trim=0 4.7cm 5.2cm 0]{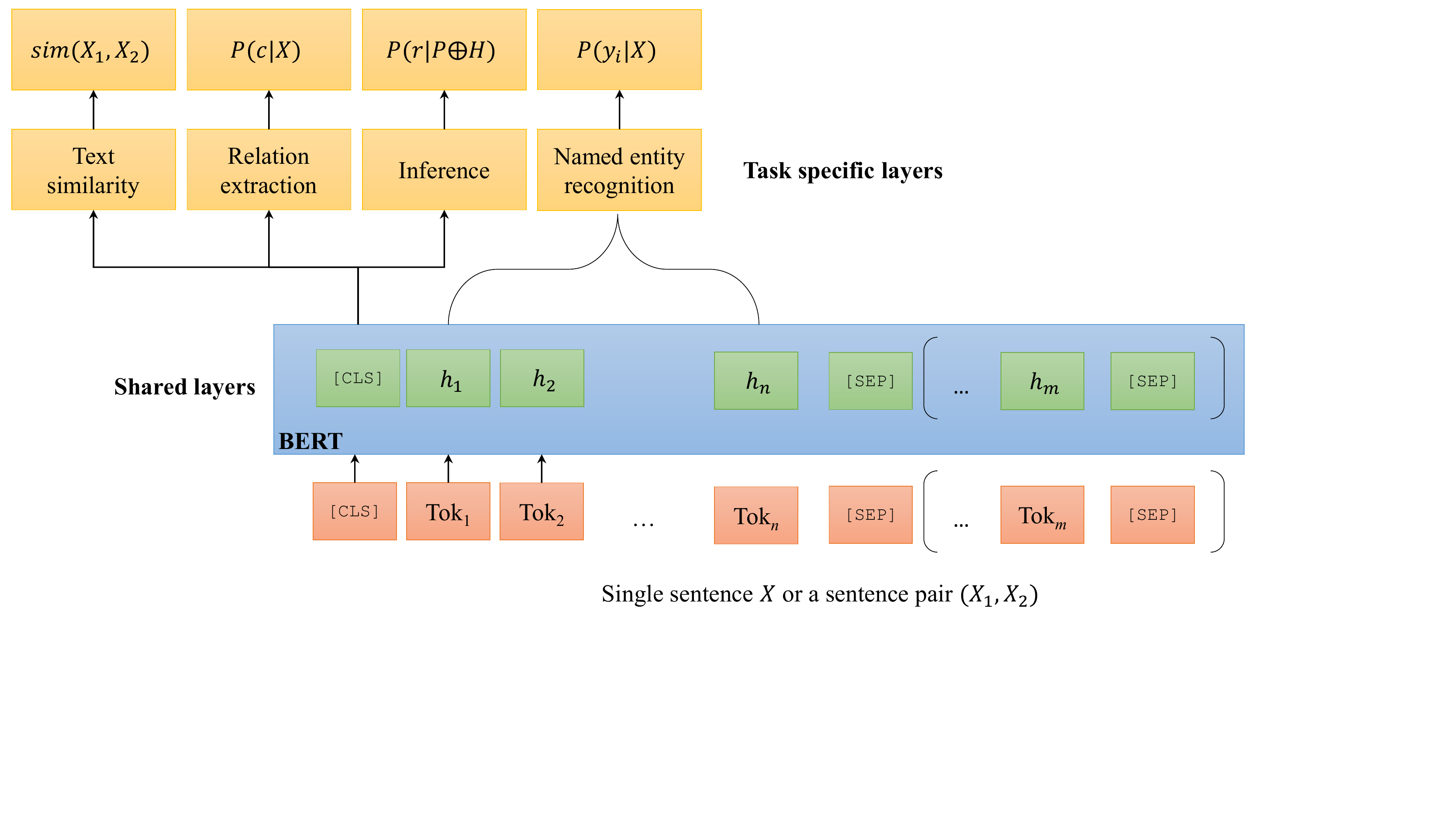}
\caption{The architecture of the MT-BERT model.}
\label{fig:architecture}
\end{figure*}

The architecture of the \MTBERT model is shown in Figure~\ref{fig:architecture}. The shared layers are based on BERT~\cite{devlin2018bert}. The input $X$ can be either a sentence or a pair of sentences packed together by a special token \verb|[SEP]|. If $X$ is longer than the allowed maximum length (e.g., 128 tokens in the BERT's base configuration), we truncate $X$ to the maximum length. When $X$ is packed by a sequence pair, we truncate the longer sequence one token at a time. Similar to~\cite{devlin2018bert}, two additional tokens are added at the start (\verb|[CLS]|) and end (\verb|[SEP]|) of $X$, respectively. Similar to~\cite{lee2020biobert, peng2019transfer}, in the sequence tagging tasks, we split one sentence into several sub-sentences if it is longer than 30 words.

In the shared layers, the BERT model first converts the input sequence to a sequence of embedding vectors. Then, it applies attention mechanisms to gather contextual information. This semantic representation is shared across all tasks and is trained by our multi-task objectives. Finally, the BERT model encodes that information in a vector for each token $(h_0,\ldots,h_n)$. 

On top of the shared BERT layers, the task-specific layer uses a fully-connected layer for each task. We fine-tune the BERT model and the task-specific layers using multi-task objectives during the training phase. More details of the multi-task objectives in the BLUE benchmark are described below.

\subsection{Sentence similarity}

Suppose that $h_0$ is the BERT's output of the token \verb|[CLS]| in the input sentence pair $(X_1,X_2)$. We use a fully connected layer to compute the similarity score $sim(X_1,X_2)=ah_0+b$, where $sim(X_1,X_2)$ is a real value. This task is trained using the Mean Squared Error (MSE) loss: $(y-sim(X_1,X_2))^2$, where $y$ is the real-value similarity score of the sentence pair.

\subsection{Relation extraction}

This task extracts binary relations (two arguments) from sentences. After replacing two arguments of interest in the sentence with pre-defined tags (e.g., GENE, or DRUG), this task can be treated as a classification problem of a single sentence $X$. Suppose that $h_0$ is the output embedding of the token \verb|[CLS]|, the probability that a relation is labeled as class $c$ is predicted by a fully connected layer and a logistic regression with softmax: $P(c \vert X)=softmax(ah_0+b)$. This approach is widely used in the transformer-based models~\cite{devlin2018bert, peng2019transfer, liu2019multi}. This task is trained using the categorical cross-entropy loss: $-\sum_c{\delta(y_c=\hat{y})\log(P(c\vert X))}$, where $\delta(y_c=\hat{y})=1$ if the classification $\hat{y}$ of $X$ is the correct ground-truth for the class $c\in C$; otherwise $\delta(y_c=\hat{y})=0$.

\subsection{Inference}

After packing the pair of premise sentences with hypothesis into one sequence, this task can also be treated as a single sentence classification problem. The aim is to find logical relation $R$ between premise $P$ and hypothesis $H$. Suppose that that $h_0$ is the output embedding of the token \verb|[CLS]| in $X=P\oplus H$, $P(R\vert P\oplus H)=softmax(ah_0+b)$. This task is trained using the categorical cross-entropy loss as above.

\subsection{Named entity recognition}

The output of the BERT model produces a feature vector sequence $\{h_i\}_{i=0}^n$ with the same length as the input sequence $X$. The MTL model predicts the label sequence by using a softmax output layer, which scales the output for a label $l\in \{1,2,\dots,L\}$ as follows: $P(\hat{y}_i=j\vert x)=\frac{\exp(h_iW_j)}{\sum_{l=1}^L{\exp(h_iW_j)}}$, where $L$ is the total number of tags. This task is trained using the categorical cross-entropy loss: $-\sum_i{\sum_{y_i}{\delta(y_i=\hat{y}_i)\log{P(y_i \vert X)}}}$.

\subsection{The training procedure}

The training procedure for \MTBERT consists of three stages: (1) pretraining the BERT model, (2) refining it via multi-task learning (\MTBERTR), and (3) fine-tuning the model using the task-specific data (\MTBERTFT).
\begin{table*}[!ht]
\centering
\begin{tabular}{llllrrr}
\toprule
Corpus & Task & Metrics & Domain & Train & Dev & Test\\
\midrule
ClinicalSTS     & Sentence similarity & Pearson  & Clinical   &    675 &    75 &   318\\
ShARe/CLEFE     & NER                 & F1       & Clinical   &  4,628 & 1,075 & 5,195\\
i2b2 2010       & Relation extraction & F1       & Clinical   &  3,110 &    11 & 6,293\\
MedNLI          & Inference           & Accuracy & Clinical   & 11,232 & 1,395 & 1,422\\
BC5CDR disease  & NER                 & F1       & Biomedical &  4,182 & 4,244 & 4,424\\
BC5CDR chemical & NER                 & F1       & Biomedical &  5,203 & 5,347 & 5,385\\
DDI             & Relation extraction & F1       & Biomedical &  2,937 & 1,004 &   979\\
ChemProt        & Relation extraction & F1       & Biomedical &  4,154 & 2,416 & 3,458\\
\bottomrule
\end{tabular}
    \caption{Summary of eight tasks in the BLUE benchmark. More details can be found in~\cite{peng2019transfer}.}
    \label{tab:tasks}
\end{table*}

\subsubsection{Pretraining}

The pretraining stage follows that of the BERT using the masked language modeling technique~\cite{devlin2018bert}. Here we used the base version. The maximum length of the input sequences is thus 128.

\subsubsection{Refining via Multi-task learning}

In this step, we refine all layers in the model. Algorithm~\ref{algo:mtdnn} demonstrates the process of multi-task learning~\cite{liu2019multi}. We first initialize the shared layers with the pre-trained BERT model and randomly initialize the task-specific layer parameters. Then we create the dataset by merging mini-batches of all the datasets. In each epoch, we randomly select a mini-batch $b_t$  of task $t$ from all datasets $D$. Then we update the model according to the task-specific objective of the task $t$. Same as in~\cite{liu2019multi}, we use the mini-batch based stochastic gradient descent to learn the parameters. 
\begin{algorithm}
\SetKwFor{Initialize}{Initialize}{}{end}
\Initialize{model parameters $\theta$}{
Shared layer parameters by BERT\;
Task-specific layer parameters randomly\;
}
Create $D$ by merging mini-batches for each dataset\;
\For{$epoch$ in $1,2,...,epoch_{max}$}{
     Shuffle $D$\;
     \For{$b_t$ in D}{
     	Compute loss: $L(\theta)$ based on task $t$\;
     	Compute gradient: $\nabla(\theta)$ \\
     	Update model: $\theta = \theta - \eta \nabla(\theta)$ \\
     }
 }
\caption{\label{algo:mtdnn} Multi-task learning.}
\end{algorithm}

\subsubsection{Fine-tuning MT-BERT}

We fine-tune existing MT-BERT that are trained in the previous stage by continue training all layers on each specific task. Provided that the dataset is not drastically different in context to other datasets, the MT-BERT model will already have learned general features that are relevant to a specific problem. Specifically, we truncate the last layer (softmax and linear layers) of the MT-BERT and replace it with a new one, then we use a smaller learning rate to train the network. 
\begin{table*}[!ht]
\setlength{\tabcolsep}{4pt}
\centering
\begin{tabular}{lccccc}
\toprule
Model           & ClinicalSTS & i2b2 2010 re & MedNLI & ShARe/CLEFE & \textit{Avg} \\ 
\midrule
\BluebertC      & 0.848 & 0.764 & 0.840 & 0.771 & \textit{0.806} \\ 
\MTBluebertCR   & 0.822 & 0.745 & 0.835 & 0.826 & \textit{0.807} \\ 
\MTBluebertCFT  & 0.840 & 0.760 & \textbf{0.846} & \textbf{0.831} & \textbf{\textit{0.819}} \\ 
\bottomrule
\end{tabular}
\caption{Test results on clinical tasks.\label{tab:test clinical}}
\vspace*{1em}
\end{table*}
\begin{table*}[!ht]
\centering
\begin{tabular}{lccccc}
\toprule
\multirow{2}{*}{Model} & \multirow{2}{*}{ChemProt} & \multirow{2}{*}{DDI} & BC5CDR & BC5CDR & \multirow{2}{*}{\textit{Avg}} \\
& & & disease & chemical\\
\midrule
\BluebertB     & 0.725 & 0.739 & 0.866 & 0.935 & \textit{0.816} \\
\MTBluebertBR  & 0.714 & 0.792 & 0.824 & 0.930 & \textit{0.815} \\
\MTBluebertBFT & \textbf{0.729} & \textbf{0.820} & 0.865 & 0.931 & \textbf{\textit{0.836}} \\ 
\bottomrule
\end{tabular}
\caption{Test results on biomedical tasks.\label{tab:test biomedical}}
\end{table*}

\section{Experiments}
\label{sec:exp}

We evaluate the proposed MT-BERT on 8 tasks in BLUE benchmarks. We compare three types of models: (1) existing start-of-the-art BERT models fine-tuned directly on each task, respectively;  (2) refinement MT-BERT with multi-task training (\MTBERTR); and (3) MT-BERT with fine-tuning (\MTBERTFT).

\subsection{Datasets}

We evaluate the performance of the models on 8 datasets in the BLUE benchmark used by~\cite{peng2019transfer}. Table~\ref{tab:tasks} gives a summary of these datasets. Briefly, ClinicalSTS is a corpus of sentence pairs selected from Mayo Clinics's clinical data warehouse ~\cite{wang2018medsts}. The i2b2 2010 dataset was collected from three different hospitals and was annotated by medical practitioners for eight types of relations between problems and treatments~\cite{uzuner20112010}. MedNLI is a collection of sentence pairs selected from \mbox{MIMIC-III}~\cite{shivade2017mednli}. For a fair comparison, we use the same training, development and test sets to train and evaluate the models. ShARe/CLEF is a collection of 299 de-identified clinical free-text notes from the \mbox{MIMIC-II} database~\cite{suominen2013overview}. This corpus is for disease entity recognition.

In the biomedical domain, the ChemProt consists of 1,820 PubMed abstracts with chemical-protein interactions~\cite{krallinger2017overview}. The DDI corpus is a collection of 792 texts selected from the DrugBank database and other 233 Medline abstracts~\cite{herrero-zazo2013ddi}. These two datasets were used in the relation extraction task for various types of relations. BC5CDR is a collection of 1,500 PubMed titles and abstracts selected from the CTD-Pfizer corpus and was used in the named entity recognition task for chemical and disease entities~\cite{li2016biocreative}. 
%

\subsection{Training}

Our implementation of MT-BERT is based on the work of~\cite{liu2019multi}.\footnote{\url{https://github.com/namisan/mt-dnn}} We trained the model on one NVIDIA\textregistered~V100 GPU using the PyTorch framework. We used the Adamax optimizer~\cite{kingma2015adam} with a learning rate of $5e^{-5}$, a batch size of 32, a linear learning rate decay schedule with warm-up over 0.1, and a weight decay of 0.01 applied to every epoch of training by following~\cite{liu2019multi}. We use the BioBERT~\cite{lee2020biobert}, BlueBERT base model~\cite{peng2019transfer}, and ClinicalBERT~\cite{alsentzer2019publicly} as the domain-specific language model\footnote{\url{https://github.com/ncbi-nlp/bluebert}}. As a result, all the tokenized texts using wordpieces were chopped to spans no longer than 128 tokens. We set the maximum number of epochs to 100.  We also set the dropout rate of all the task-specific layers as 0.1.  To avoid the exploding gradient problem, we clipped the gradient norm within 1. To fine-tune the MT-BERT on specific tasks, we set the maximum number of epochs to 10 and learning rate $e^{-5}$.
\begin{figure*}[ht]
\centering
\includegraphics[width=.99\textwidth,clip,trim=0 9cm 2cm 0]{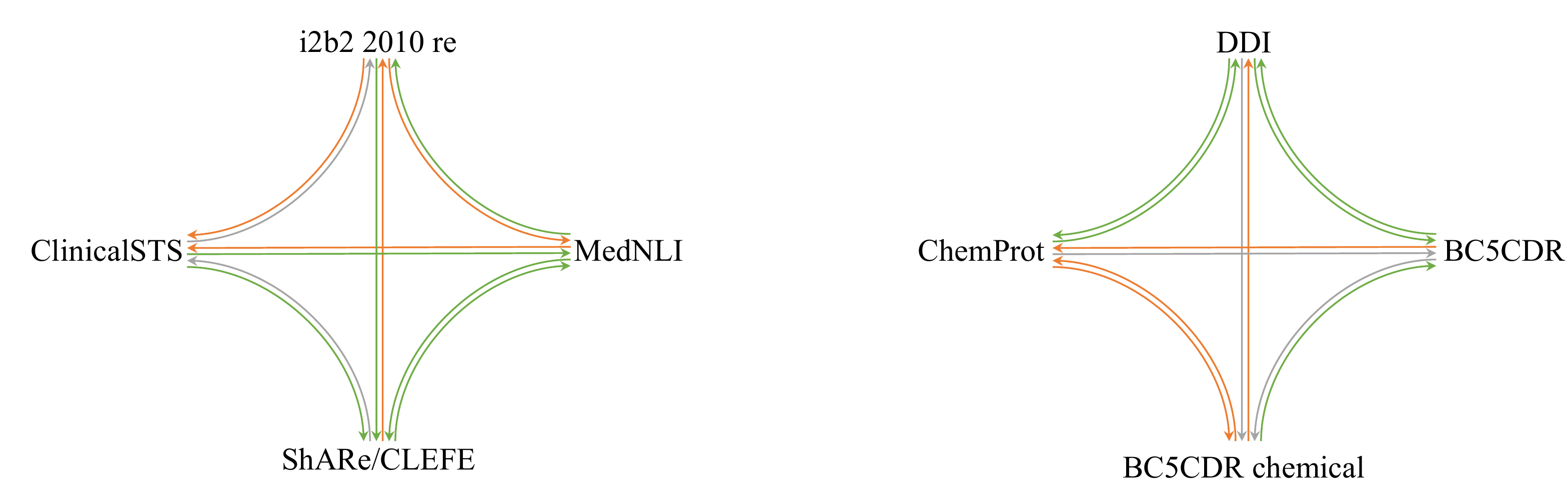}
\caption{Pairwise MTL relationships in clinical (left) and biomedical (right) domains.}
\label{fig:pairwise}
\vspace*{1em}
\end{figure*}
\begin{table*}[!ht]
\centering
\setlength{\tabcolsep}{4pt}
\begin{tabular}{lccccc@{}}
\toprule
Model & ClinicalSTS & i2b2 2010 re & MedNLI & ShARe/CLEFE & \textit{Avg}\\
\midrule
\MTClinicalBERTFT & 0.816 & 0.746 & 0.834 & 0.817 & \textit{0.803}\\ 
\MTBioBERTFT      & 0.837 & 0.741 & 0.832 & 0.818 & \textit{0.807}\\ 
\MTBluebertBFT    & 0.824 & 0.738 & 0.824 & 0.825 & \textit{0.803}\\ 
\MTBluebertCFT    & \textbf{0.840} & \textbf{0.760} & \textbf{0.846} & \textbf{0.831} & \textbf{\textit{0.819}} \\
\bottomrule
\end{tabular}
\caption{Test results of \MTBERTFT models on clinical tasks.\label{tab:test clinical fine-tune}}
\vspace*{.5em}
\end{table*}
\begin{table*}[!ht]
\centering
\begin{tabular}{lccccc}
\toprule
\multirow{2}{*}{Model} & \multirow{2}{*}{ChemProt} & \multirow{2}{*}{DDI} & BC5CDR & BC5CDR & \multirow{2}{*}{\textit{Avg}} \\
& & & disease & chemical\\
\midrule
\MTBioBERTFT   & \textbf{0.729} & 0.812 & 0.851 & 0.928 & \textit{0.830} \\ 
\MTBluebertBFT & \textbf{0.729} & \textbf{0.820} & \textbf{0.865} & \textbf{0.931} & \textbf{\textit{0.836}} \\ 
\MTBluebertCFT & 0.714 & 0.792 & 0.824 & 0.930 & \textit{0.815} \\ \bottomrule
\end{tabular}
\caption{Test results of \MTBERTFT models on biomedical tasks.\label{tab:test biomedical fine-tune}}
\end{table*}

\subsection{Results}

One of the most important criteria of building practical systems is fast adaptation to new domains. To evaluate the models on different domains, we multi-task learned various MT-BERT on BLUE biomedical tasks and clinical tasks, respectively. \BluebertC is the base BlueBERT model pretrained on PubMed abstracts and MIMIC-III clinical notes, and fine-tuned for each BLUE task on task-specific data. MT- model are the proposed models described in Section~\ref{sec:multitask}. We used the pre-trained \BluebertC to initialize its shared layers, refined the model via MTL on the BLUE tasks (\MTBluebertCR). We keep fine-tuning the model for each BLUE task using task-specific data, then got \MTBluebertCFT. 

Table~\ref{tab:test clinical} shows the results on clinical tasks. \MTBluebertCFT created new state-of-the-art results on 2 tasks and pushing the benchmark to 81.9\%, which amounts to 1.3\% absolution improvement over \BluebertC and 1.2\% absolute improvement over \MTBluebertCR. On the ShAReCLEFE task, the model gained the largest improvement by 6\%. On the MedNLI task, the MT model gained improvement by 2.4\%. On the remaining tasks, the MT model also performed well by reaching the state-of-the-art performance with less than 1\% differences. When compared the models with and without fine-tuning on single datasets, Table~\ref{tab:test clinical} shows that the multi-task refinement model is similar to single baselines on average. Consider that \MTBluebertCR is one model while \BluebertC are 4 individual models, we believe the MT refinement model would bring the benefit when researchers are in the hassle of choosing the suitable model for new problems or problems with limited training data.

In biomedical tasks, we used \BluebertB as the baseline because it achieved the best performance on the BLUE benchmark. Table~\ref{tab:test biomedical} shows the similar results as in the clinical tasks. \MTBluebertBFT created new state-of-the-art results on 2 tasks and pushing the benchmark to 83.6\%, which amounts to 2.0\% absolute improvement over \BluebertB and 2.1\% absolute improvement over \MTBluebertBR. On the DDI task, the model gained the largest improvement by 8.1\%.


\subsection{Discussion}
\begin{table*}[!ht]
\setlength{\tabcolsep}{3pt}
\centering
\begin{tabular}{lccccc}
\toprule
\multirow{2}{*}{Model} & BlueBERT & BlueBERT & MT-BioBERT & MT-BlueBERT & MT-BlueBERT \\
& \textsubscript{biomedical} & \textsubscript{clinical} & Fine-Tune & Fine-Tune\textsubscript{biomedical} & Fine-Tune\textsubscript{clinical} \\
\midrule
ClinicalSTS     & 0.845 & \textbf{0.848} & 0.807 & 0.820 & 0.807\\
i2b2 2010 re    & 0.744 & \textbf{0.764} & 0.740 & 0.738 & 0.748\\
MedNLI          & 0.822 & 0.840 & 0.831 & 0.814 & \textbf{0.842}\\
ChemProt        & 0.725 & 0.692 & \textbf{0.735} & 0.724 & 0.686\\
DDI             & 0.739 & 0.760 & \textbf{0.810} & 0.808 & 0.779\\
BC5CDR disease  & \textbf{0.866} & 0.854 & 0.849 & 0.853 & 0.848\\
BC5CDR chemical & \textbf{0.935} & 0.924 & 0.928 & 0.928 & 0.914\\
ShARe/CLEFE     & 0.754 & 0.771 & 0.812 & 0.814 & \textbf{0.830}\\
\midrule
\textit{Avg}    & \textit{0.804} & \textit{0.807} & \textbf{\textit{0.814}} & \textit{0.812} & \textit{0.807}\\
\bottomrule
\end{tabular}
\caption{Test results on eight BLUE tasks.\label{tab:test blue}}
\end{table*}

\subsubsection{Pairwise MTL}

To investigate which tasks are beneficial or harmful to others, we train on two tasks jointly using \MTBluebertBR and \MTBluebertCR. Figure~\ref{fig:pairwise} gives pairwise relationships. The directed green (or red and grey) edge from $s$ to $t$ means $s$ improves (or decreases and has no effect on) $t$. 

In the clinical tasks, ShARe/CLEFE always gets benefits from multi-task learning the remaining 3 tasks as the incoming edges are green. One factor might be that ShARe/CLEFE is an NER task that generally requires more training data to fulfill the data-hungry need of the BERT model. ClinicalSTS helps MedNLI because the nature of both are related and their inputs are a pair of sentences. MedNLI can help other tasks except ClinicalSTS partially because the test set of ClinialSTS is too small to reflect the changes. We also note that i2b2 2010 re can be both beneficial and harmful, depending on which other tasks they are trained with. One potential cause is i2b2 2010 re was collected from three different hospitals and have the largest label size of 8. 

In the biomedical tasks, both DDI and ChemProt tasks can be improved by MTL on other tasks, potentially because they are harder with largest size of label thus require more training data. In the meanwhile, BC5CDR chemical and disease can barely be improved potentially because they have already got large dataset to fit the model. 

\subsubsection{MTL on BERT variants}

First, we would like to compare multi-task learning on BERT variants: BioBERT, ClinicalBERT, and BlueBERT. In the clinical tasks (Table~\ref{tab:test clinical fine-tune}), \MTBluebertCFT outperforms other models on all tasks. When compared the MTL models using BERT model pretrained on PubMed only (rows 2 and 3) and on the combination of PubMed and clinical notes (row 4), it shows the impact of using clinical notes during the pretraining process. This observation is consistently as shown in~\cite{peng2019transfer}. On the other hand, \MTClinicalBERTFT, which used ClinicalBERT during the pretraining, drops $\sim$1.6\% across the tasks. The differences between ClinicalBERT and BlueBERT are at least in 2-fold. (1) ClinicalBERT used ``cased'' text while BlueBERT used ``uncased'' text; and (2) the number of epochs to continuously pretrained the model. Given that there are limited details of pretraining ClinicalBERT, further investigation may be necessary. 

In the biomedical tasks, Table~\ref{tab:test biomedical fine-tune} shows that \MTBioBERTFT and \MTBluebertBFT reached comparable results and pretraining on clinical notes has a negligible impact.

\subsubsection{Results on all BLUE tasks}

Next, we also compare MT-BERT with its variants on all BLUE tasks. Table~\ref{tab:test blue} shows that \MTBioBERTFT reached the best performance on average and \MTBluebertBFT stays closely. While confusing results were obtained when combing variety of tasks in both biomedical and clinical domains, we observed again that MTL models pretrained on biomedical literature perform better in biomedical tasks; and MTL models pretrained on both biomedical literature and clinical notes perform better in clinical tasks. These observations may suggest that it might be helpful to train separate deep neural networks on different types of text genres in BioNLP.

\section{Conclusions and future work}

In this work, we conduct an empirical study on MTL for biomedical and clinical tasks, which so far has been mostly studied with one or two tasks. Our results provide insights regarding domain adaptation and show benefits of the MTL refinement and fine-tuning. We recommend a combination of the MTL refinement and task-specific fine-tuning approach based on the evaluation results. When learned and fine-tuned on a different domain, MT-BERT achieved improvements by 2.0\% and 1.3\% in biomedical and clinical domains, respectively. Specifically, it has brought significant improvements in 4 tasks.

There are two limitations to this work. First, our results on MTL training across all BLUE benchmark show that MTL is not always effective. We are interested in exploring further the characterization of task relationships. For example, it is not clear whether there are data characteristics that help to determine its success~\cite{martinezalonso2017when, changpinyo2018multi}. In addition, our results suggest that the model could benefit more from some specific examples of some of the tasks in Table~\ref{tab:tasks}. For example, it might be of interest to not using the BC5CDR corpus in the relation extraction task in future. Second, we studied one approach to MTL by sharing the encoder between all tasks while keeping several task-specific decoders. Other approaches, such as fine-tuning only the task specific layers, soft parameter sharing~\cite{ruder2017overview}, knowledge distillation~\cite{liu2019improving}, need to be investigated in the future.

While our work only scratches the surface of MTL in the medical domain, we hope it will shed light on the development of generalizable NLP models and task relations that can lead to gains from MTL models over single-task setups.

\section*{Acknowledgments}

This work was supported by the Intramural Research Programs of the NIH National Library of  Medicine. This work was also supported by the National Library of Medicine of the National Institutes of Health under award number K99LM013001. We are also grateful to the authors of mt-dnn (\url{https://github.com/namisan/mt-dnn}) to make the codes publicly available.

\bibliography{acl2020}
\bibliographystyle{acl_natbib}

\end{document}